%% file: main_en.tex
\documentclass[letterpaper,10pt,conference]{ieeeconf}
\usepackage[T1]{fontenc}

\usepackage{amsmath}
\usepackage{newtxmath}
\usepackage{bm}
\usepackage{graphicx}
\usepackage{booktabs}
\usepackage{multirow}
\usepackage{tabularx}
\usepackage{array}
\usepackage{cite}
\usepackage{xcolor}
\usepackage{colortbl}
\usepackage{url}
\usepackage{flushend}
\usepackage{placeins}
\usepackage{tikz}
\usetikzlibrary{positioning,arrows.meta,fit,calc}
\usepackage[protrusion=false]{microtype}

\newcommand{\ours}{\textsc{TIO-Former}}
\newcommand{\range}{\mathbf D}
\newcommand{\imu}{\mathbf I}
\newcommand{\SE}{\mathrm{SE}}

\title{\LARGE\bfseries TIO-Former: Ultra-Lightweight 6-Directional ToF-Inertial Odometry for Nano-UAVs via a Streaming Causal Transformer}
\author{Yang Liu\textsuperscript{\textdagger}\textsuperscript{\rm 1,2},
Yifan He\textsuperscript{\textdagger}\textsuperscript{\rm 2},
Wenhao Zhao\textsuperscript{\textdagger}\textsuperscript{\rm 2},
Xiangyu Mo\textsuperscript{\rm 2},
Yang Xu\textsuperscript{\rm 2},
Hao Wei\textsuperscript{\rm 2},\\
Mingze Ma\textsuperscript{\rm 2},
Huan Li\textsuperscript{\rm 2},
Yifan Wu\textsuperscript{\rm 2},
Fei Gao\textsuperscript{\rm 1,2},
Zipeng Dai\textsuperscript{\rm 1,2},
Xin Zhou\textsuperscript{*}\textsuperscript{\rm 1,2}\\[0.5ex]
{\normalsize \textsuperscript{\rm 1}Zhejiang University, Hangzhou, China}\\
{\normalsize \textsuperscript{\rm 2}Differential Robotics, Hangzhou, China}\\
{\normalsize \textsuperscript{\textdagger}These authors contributed equally to this work.}\\
{\normalsize \textsuperscript{*}Corresponding author. E-mail: 923137104@qq.com (Xin Zhou).}}

\begin{document}
\maketitle
\thispagestyle{empty}
\pagestyle{empty}

\input{sections_en/abstract}
\input{sections_en/introduction}
\input{sections_en/related_work}
\input{sections_en/method}
\input{sections_en/experiments}
\FloatBarrier
\input{sections_en/conclusion}
\bibliographystyle{IEEEtran}
\bibliography{references}
\end{document}

%% file: sections_en/abstract.tex
\begin{abstract}
Autonomous nano-UAV navigation requires accurate ego-motion estimation under stringent size, weight, power, and computing (SWaP-C) constraints, where visual sensors and LiDARs exceed payload limits, optical flow degrades in low-texture scenes, and inertial-only state estimation is susceptible to accumulated drift. While multi-zone time-of-flight (ToF) arrays provide a lightweight metric complement, 6-DoF estimation from merely 384 ranges per frame is challenged by invalid returns, anisotropic observability, and temporal computational scaling. We propose \ours{}, a camera-free, optical-flow-free, and mapless range-inertial odometry framework driven by an IMU and an ultra-lightweight ($15$~g) payload of six orthogonal $8\times8$ ToF arrays. Our frontend pairs consecutive range grids with a bilateral gated difference, while IMU-guided cross-attention dynamically routes directional features conditioned on platform kinematics. A Streaming Causal Transformer couples an uncompressed Local KV cache with compressed Chunk-FIFO memory, maintaining bounded inference cost and memory footprint independent of flight duration. In real-flight evaluations, \ours{} reduces open-loop position error by $54.4\%$ compared to nano-UAV optical flow and by $66.4\%$--$89.1\%$ over learned inertial baselines. We also evaluate performance across multiple environments and robustness under severe sensing degradation. Deployed on an edge RISC-V companion computer, \ours{} achieves a P95 latency of $10.466$~ms and peak resident memory of $6.324$~MiB ($<5\%$ system RAM), demonstrating that sparse range sensing provides practical geometric anchoring for resource-constrained micro-aerial robots. Code is available at \url{https://github.com/Ly041021/TIO-Former}.
\end{abstract}

%% file: sections_en/introduction.tex
\section{Introduction}
\label{sec:introduction}

Deploying autonomous nano-scale unmanned aerial vehicles (nano-UAVs) in confined, GPS-denied environments demands accurate and drift-resilient ego-motion estimation under stringent size, weight, power, and computing (SWaP-C) budgets. While visual-inertial odometry (VIO) and LiDAR SLAM provide rich geometric constraints~\cite{qin2018vinsmono,campos2021orbslam3}, their physical payload and computational burdens remain prohibitive for gram-scale platforms. Downward optical flow offers a lightweight alternative~\cite{mcguire2017pocketdrone}, but degrades severely over low-texture surfaces or in dim lighting; range-based height estimates can also become unreliable over uneven terrain. Inertial-only state estimation operates independently of visual appearance but remains susceptible to accumulated drift. Multi-zone time-of-flight (ToF) arrays present an attractive metric alternative: arranged orthogonally, six $8\times8$ sensors supply 384 direct range measurements per frame with negligible bandwidth overhead.

\begin{figure}[!t]
\centering
\includegraphics[width=\columnwidth]{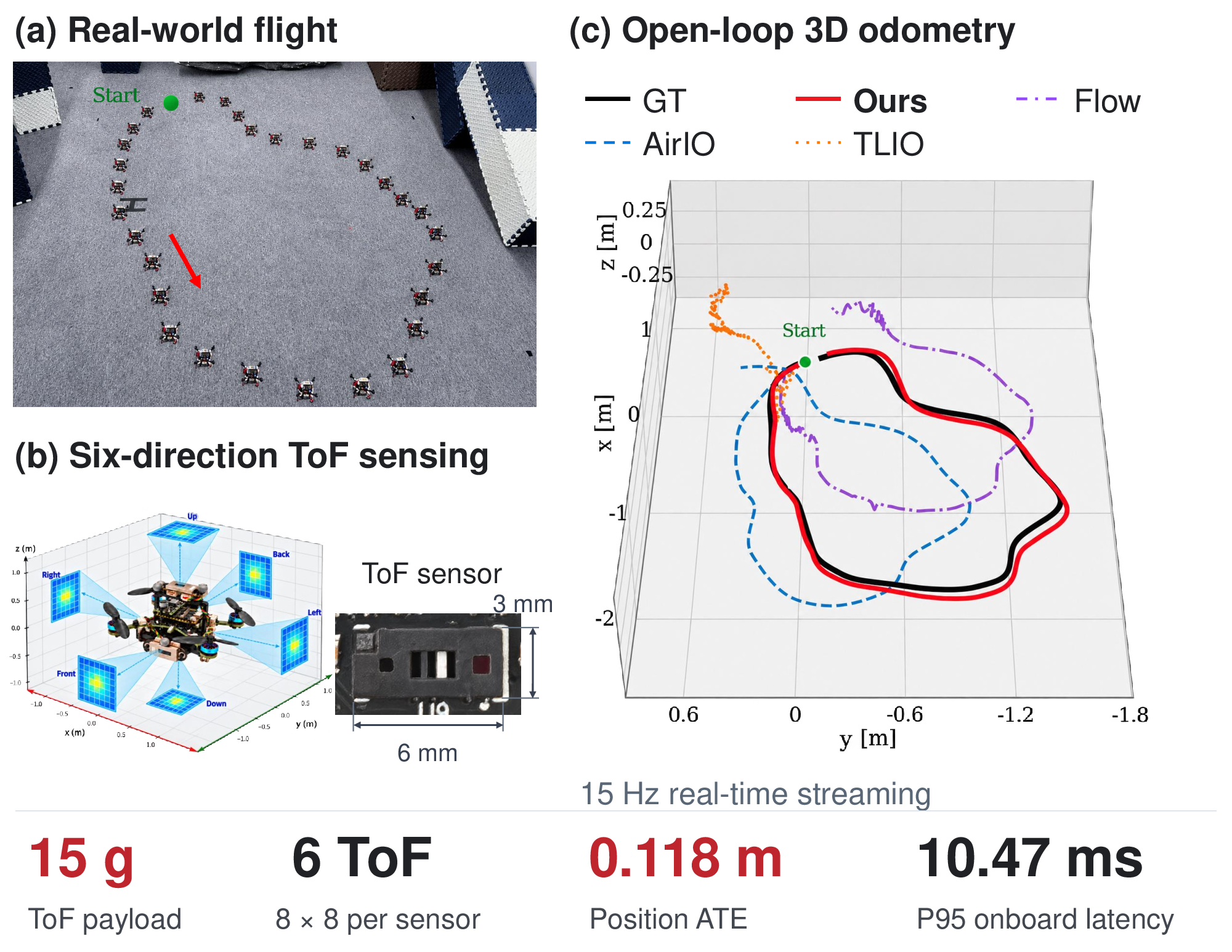}
\caption{Overview of \ours{}. (a) Real-world flight. (b) Six-direction ToF sensing with six $8\times8$ arrays and a 15~g ToF payload. (c) Open-loop 3D trajectory comparison against ground truth (GT), Crazyflie Flow, AirIO, and TLIO. The bottom row summarizes position ATE and onboard latency; streaming inference runs at 15~Hz.}
\label{fig:teaser}
\end{figure}

Prior sparse-ToF navigation has largely targeted reactive obstacle avoidance, pre-mapped localization, or planar SLAM assisted by optical flow~\cite{muller2023matrixtof,muller2023toflocalization,friess2024tofslam}. In this paper, we show that such ultra-sparse range sensing can support continuous, standalone 6-DoF range-inertial odometry without cameras, optical flow, or prior maps. Realizing this capability requires addressing three coupled challenges: (i) measurement dropouts and invalid returns introduce spurious inter-frame range transitions; (ii) directional observability varies dynamically with platform motion and scene layout; and (iii) maintaining temporal context over continuous flights quickly breaches onboard memory and latency bounds.

Simple zero-padding or heuristic imputation of missing returns creates artificial motion discontinuities across validity transitions. Furthermore, sparse range observations constrain spatial motion axes anisotropically, making uniform feature pooling suboptimal and demanding motion-conditioned geometric fusion. While temporal context is critical to resolve geometric degeneracies, storing an uncompressed key--value history scales linearly with flight duration. A viable estimator must therefore explicitly gate measurement validity, dynamically route directional observations using inertial context, and retain recent temporal detail and compressed earlier context within fixed per-step compute and memory budgets.

To address these challenges, we propose \ours{} (Fig.~\ref{fig:teaser}), an end-to-end learned range-inertial odometry framework. First, consecutive range grids are coupled with a bilateral gated difference, which weights inter-frame range changes by measurement reliability at both temporal endpoints, retaining absolute metric depth while suppressing artifacts from missing returns. Second, inertial features extracted over a short temporal window dynamically query directional ToF tokens through cross-attention, prioritizing geometrically informative views conditioned on the instantaneous motion state. Third, a Streaming Causal Transformer pairs an uncompressed Local KV cache with bounded Chunk-FIFO memory, maintaining bounded per-step inference cost and memory footprint independent of trajectory duration. Finally, multi-horizon trajectory supervision reinforces integrated trajectory consistency on the $SE(3)$ manifold alongside step-wise relative pose regression.

Our primary contributions are summarized as follows:
\begin{itemize}
    \item A camera-free, optical-flow-free, and mapless 6-DoF range-inertial odometry framework driven solely by an IMU and six orthogonal $8\times8$ ToF arrays with a total ToF sensor mass of $15$~g, enabling continuous metric localization.
    \item A reliability-gated range frontend and an IMU-guided directional cross-attention module, coupled with a two-tier streaming Transformer that maintains bounded per-step inference cost and memory footprint independent of flight duration.
    \item Full-stack onboard deployment on a nano-UAV and evaluation across a real-flight benchmark ($>4.5$~km over four distinct indoor scenes), demonstrating real-time edge execution on a low-power RISC-V SoC while significantly outperforming commercial optical flow and learned inertial baselines.
\end{itemize}

%% file: sections_en/related_work.tex
\section{Related Work}
\label{sec:related_work}

\subsection{Lightweight Perception and Sparse Range Sensing}
Visual-inertial odometry (VIO) and LiDAR SLAM deliver robust metric state estimation~\cite{qin2018vinsmono,campos2021orbslam3}, yet their sensing payloads and computational demands exceed the SWaP-C envelope of nano-UAVs. Downward optical flow enables low-power velocity estimation~\cite{mcguire2017pocketdrone}, and PULP-DroNet demonstrates microcontroller-class visual navigation~\cite{palossi2019pulpdronet}; however, vision-based cues degrade drastically under aggressive maneuvers, motion blur, or poor illumination. Multi-zone time-of-flight (ToF) arrays provide direct metric range constraints at ultra-low spatial resolution. Prior studies utilize matrix ToF primarily for reactive obstacle avoidance~\cite{muller2023matrixtof}, localization against pre-built maps~\cite{muller2023toflocalization}, or planar swarm SLAM aided by optical flow~\cite{friess2024tofslam}. While dense range-inertial systems have been explored~\cite{koide2025rangeinertial}, they depend on bulky, high-bandwidth LiDAR sensors. In contrast, our work achieves standalone 6-DoF relative state estimation without cameras, optical flow, or prior maps, operating exclusively on six orthogonal $8\times8$ ToF arrays and a single IMU.

\subsection{Learned Odometry and Temporal Fusion}
Learned inertial odometry models regress relative displacements or synthetic kinematic measurements to mitigate integration drift~\cite{chen2018ionet,herath2020ronin,liu2020tlio}. Recent efforts improve IMU feature observability~\cite{qiu2025airio}, incorporate aerodynamic priors via Transformers and extended Kalman filters~\cite{cui2026aiio}, optimize microcontroller deployment~\cite{saha2022tinyodom}, and learn transferable inertial representations~\cite{zhao2025tartanimu}. While these approaches rely solely on proprioceptive sensing, our framework introduces sparse exteroceptive range measurements to continuously anchor dead reckoning against external geometry.

Multimodal learning frameworks fuse heterogeneous streams to improve estimation robustness~\cite{clark2017vinet,chen2019selectivefusion,lu2020milliego}, frequently utilizing cross-attention for inter-modal alignment~\cite{wei2025botvio} and causal attention for temporal tracking~\cite{kurt2024causalvio}. Rather than adopting generic feature concatenation, we formulate directional cross-attention where inertial motion queries selectively retrieve spatial tokens across six discrete physical directions, dynamically down-weighting degraded or uninformative views.

Recurrent networks summarize history in a fixed-size state, whereas unrestricted causal Transformer caches grow with flight duration. Sliding windows bound this growth but discard older tokens. Our two-tier memory combines an uncompressed Local KV cache with fixed-capacity Chunk-FIFO memory of learned summaries, preserving recent detail and compressed earlier context within bounded inference and memory budgets.

%% file: sections_en/method.tex
\section{Methodology}
\label{sec:method}

\begin{figure*}[t]
\centering
\includegraphics[width=\textwidth]{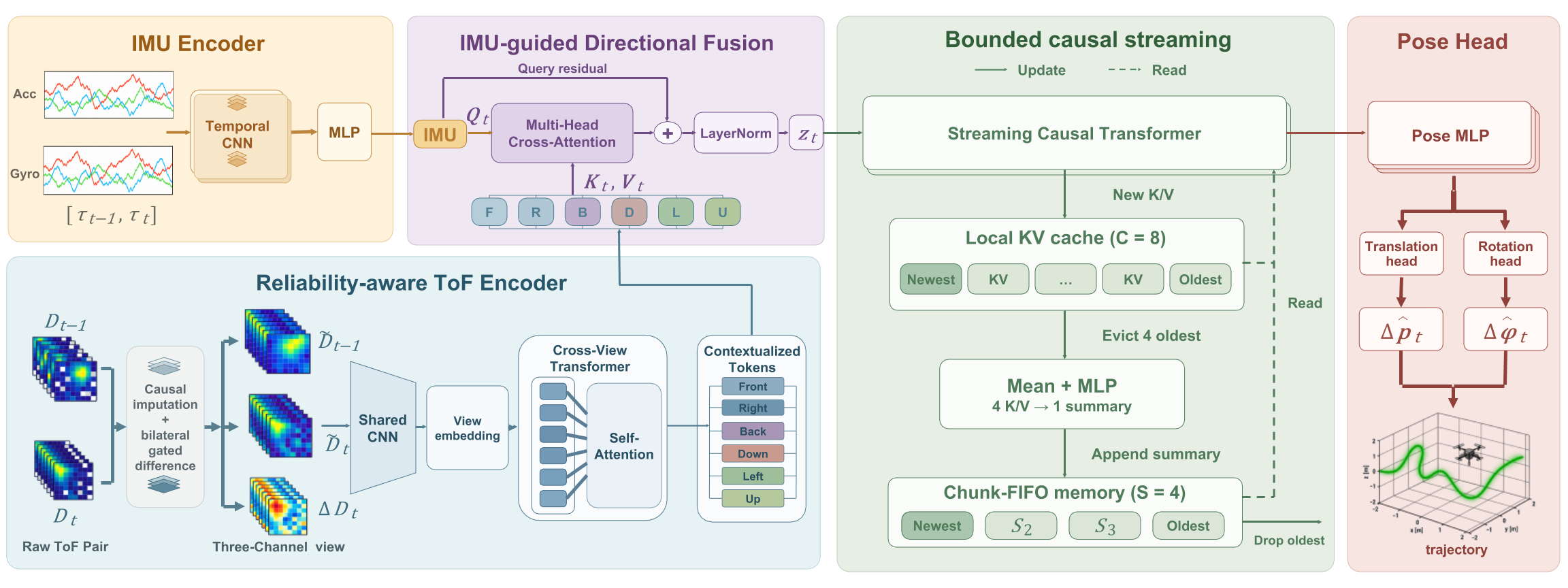}
\caption{Architecture of \ours. Causal imputation and the bilateral gated difference feed a weight-shared CNN and a cross-view Transformer. IMU queries then aggregate the six directional ToF tokens, while a Streaming Causal Transformer combines a Local KV cache with compressed Chunk-FIFO memory for bounded streaming inference.}
\label{fig:architecture}
\end{figure*}

\subsection{State and Inputs}
\label{subsec:problem_formulation}

Let $\mathbf T_t=(\mathbf R_t,\mathbf p_t)\in\SE(3)$ denote the transformation from the body frame to the world frame at time step $t$. At each $15$~Hz cycle, the system receives six orthogonal ToF grids $\range_t^{(i)}\in\mathbb R^{8\times8}$ ($i=1,\ldots,6$), collectively denoted as $\range_t$, alongside synchronized $200$~Hz IMU angular velocities and linear accelerations resampled within the interval $[\tau_{t-1},\tau_t]$ to form $\imu_t\in\mathbb R^{8\times6}$. Raw millimeter ranges are converted to meters ($d=10^{-3}d_{\mathrm{mm}}$) to align with SI inertial units, retaining absolute metric scale without artificial normalization.

The relative motion target $\Delta\mathbf T_t=\mathbf T_{t-1}^{-1}\mathbf T_t$ comprises the body-frame translation $\Delta\mathbf p_t=\mathbf R_{t-1}^{\top}(\mathbf p_t-\mathbf p_{t-1})$ and relative rotation $\Delta\mathbf R_t=\mathbf R_{t-1}^{\top}\mathbf R_t$. Given the predicted relative translation $\Delta\hat{\mathbf p}_t$ and axis-angle rotation vector $\Delta\hat{\bm\phi}_t$, the trajectory is recursively integrated as:
\begin{equation}
\begin{aligned}
\Delta\hat{\mathbf R}_t&=\exp([\Delta\hat{\bm\phi}_t]_\times),\\
\hat{\mathbf R}_t&=\hat{\mathbf R}_{t-1}\Delta\hat{\mathbf R}_t,\qquad
\hat{\mathbf p}_t=\hat{\mathbf p}_{t-1}+\hat{\mathbf R}_{t-1}\Delta\hat{\mathbf p}_t.
\end{aligned}
\label{eq:kinematics}
\end{equation}
Here $[\mathbf a]_\times\mathbf b=\mathbf a\times\mathbf b$ defines the skew-symmetric cross-product matrix, and $(\cdot)^\vee$ is its inverse map, satisfying $([\mathbf a]_\times)^\vee=\mathbf a$. The operators $\exp$ and $\log$ denote the matrix exponential and logarithm. Because body-frame translations are accumulated using the estimated orientations, orientation errors also contribute to positional drift, motivating multi-horizon trajectory supervision.

\subsection{Reliability-Aware ToF Encoding}
\label{subsec:tof_pair}

The operations below apply independently to each sensor grid; the directional index $(i)$ is omitted until inter-view fusion. Let $\tilde{\range}_t$ denote a causally imputed range grid and $\Delta\range_t$ denote the bilateral gated difference.

\textbf{Reliability and causal imputation.}
Specular reflections, multi-path interference, and out-of-range surfaces frequently cause missing returns or heavy measurement noise in multi-zone ToF sensors. Furthermore, sudden transitions between valid returns and zero/default values induce spurious apparent motion. We decouple geometric structure from measurement validity by constructing a deterministic reliability mask $\mathbf A_t\in[0,1]^{8\times8}$ based on sensor return status $s$ and measured range $d$:
\begin{equation}
w(d,s)=\mathbb I[0.015\le d\le4.0]\,
\begin{cases}
1,&s=0,\\
0.5,&s=10,\\
0,&\text{otherwise},
\end{cases}
\label{eq:quality}
\end{equation}
where the sensor status flag $s$ reflects measurement confidence: $s=0$ indicates a verified valid return, while $s=10$ denotes a newly acquired target not detected in the preceding cycle (metrically valid but lacking temporal tracking continuity). All other status codes (e.g., weak signal, excessive noise, or non-detections) indicate invalid measurements ($w=0$). Pixels with $s=10$ are assigned a reduced weight of $0.5$ and excluded as spatial interpolation donors. Invalid pixels are causally imputed using valid spatial neighbors and preceding measurements to produce $\tilde{\range}_t$, while their reliability weights remain zero to distinguish imputed estimates from true observations.

\textbf{Training perturbations.}
To improve robustness to observation dropouts and fluctuating returns, we introduce stochastic degradation masks and autoregressive noise during training. Training sequences undergo corruption with probability $0.6$: discrete, spatial-patch, or temporal-burst masks with missing ratios $\rho\in\{0.1,0.3,0.5\}$ zero out both ranges and reliability weights of $\lfloor\rho|\Omega_{\mathrm{valid}}|\rfloor$ valid pixels prior to imputation. Retained returns are perturbed with autoregressive colored noise $n_t=\gamma n_{t-1}+\epsilon_t$, where $\gamma=0.9$ and $\epsilon_t\sim\mathcal N(0,(1-\gamma^2)\sigma^2)$ with $\sigma=0.01$~m.

\textbf{Bilateral gated difference.}
Here, bilateral refers to joint gating by the reliability weights at both temporal endpoints, $t-1$ and $t$:
\begin{equation}
\begin{aligned}
\Delta\range_t&=\mathbf A_{t-1}\odot\mathbf A_t\odot(\tilde{\range}_t-\tilde{\range}_{t-1}),\\
\mathbf X_t&=\operatorname{Cat}(\tilde{\range}_{t-1},\tilde{\range}_t,\Delta\range_t)\in\mathbb R^{3\times8\times8}.
\end{aligned}
\label{eq:pair_input}
\end{equation}
\begin{figure}[!tb]
\centering
\includegraphics[width=\columnwidth]{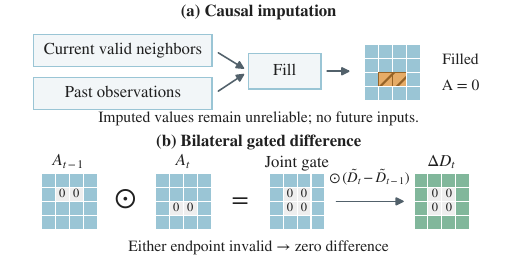}
\caption{Causal imputation and the bilateral gated difference, illustrated with schematic grid excerpts and binary reliability masks. (a) Invalid entries are filled using current valid neighbors and past observations while retaining zero reliability (hatched orange cells). (b) The element-wise product of the masks at $t-1$ and $t$ gates the range difference: an invalid endpoint yields zero, while green cells retain differences valid at both endpoints.}
\label{fig:tof_reliability}
\end{figure}
The three-channel input provides an explicit inductive bias by representing metric scene geometry and inter-frame range changes in separate channels. While consecutive grids $[\tilde{\range}_{t-1}, \tilde{\range}_t]$ anchor absolute distance to obstacle boundaries and surface orientations, requiring standard 2D spatial convolutions to implicitly deduce motion across input channels forces the network to approximate numerical differentiation over noisy and sparse measurements. Explicitly injecting $\Delta\range_t$ directly exposes first-order inter-frame range-change cues along the sensor boresight, capturing dynamic scene expansion, contraction, and tilt that correlate with platform linear velocity and angular rates.

The bilateral gated difference accounts for measurement validity at both temporal endpoints. Although causal imputation provides numerical continuity for 2D convolutions, naive differencing over filled entries would introduce severe false motion artifacts whenever measurement validity transitions occur—such as target re-acquisition or sudden dropouts. The joint endpoint weight $\mathbf A_{t-1}\odot\mathbf A_t$ suppresses a difference when either endpoint is invalid and attenuates it when either endpoint has reduced reliability. Thus, $\mathbf X_t$ provides the directional encoder with metric spatial information and reliability-weighted range-change cues.

\textbf{Spatial view encoding.}
To minimize the parameter footprint and memory bandwidth on resource-constrained hardware, a single weight-shared 2D CNN $f_D$ extracts local spatial features across all six sensor grids. Because local geometric primitives—such as surface normals, obstacle boundaries, and radial range-rate gradients—exhibit identical measurement physics regardless of mounting orientation, weight sharing provides effective cross-view statistical regularization while avoiding separate CNN parameters for each view. To subsequently restore directional identity and decouple local geometric extraction from platform extrinsic orientation, learnable physical view embeddings $\mathbf e_{\mathrm{view}}^{(i)}$ are added to the pooled representations. A cross-view Transformer layer then models omnidirectional spatial dependencies:
\begin{equation}
\begin{aligned}
\mathbf e_t^{(i)}&=f_D(\mathbf X_t^{(i)})+\mathbf e_{\mathrm{view}}^{(i)},\\
\mathbf F_t&=\operatorname{Tr}_{\mathrm{view}}([\mathbf e_t^{(1)},\ldots,\mathbf e_t^{(6)}])\in\mathbb R^{6\times d_{\mathrm{f}}},
\end{aligned}
\end{equation}
where $\mathbf F_t=[\mathbf f_t^{(1)},\ldots,\mathbf f_t^{(6)}]$ denotes the contextualized directional tokens. This omnidirectional self-attention distributes global geometric context across viewpoints before motion-conditioned selection.

\subsection{IMU-Guided Directional Fusion}
\label{subsec:crossattn}

Following AI-IO~\cite{cui2026aiio}, we deploy a three-layer 1D CNN as the inertial encoder, mapping $\imu_t$ to a local inertial feature $\mathbf h_t^I=f_I(\imu_t)$ that captures high-frequency body dynamics across the sampling window.

The motion information provided by each directional range view depends on the scene geometry and the vehicle motion. For instance, forward acceleration induces pronounced range-rate variations along the frontal boresight, whereas lateral views parallel to planar walls or open corridors yield geometrically degenerate returns. Uniform pooling assigns equal weights to all views, whereas channel concatenation does not explicitly compute motion-conditioned view weights. Motivated by these differences and prior cross-modal attention methods in visual-inertial state estimation~\cite{wei2025botvio}, we use IMU-guided cross-attention to weight directional ToF features. Using inertial features as queries, the module attends to the six directional ToF tokens: directions aligned with dominant motion axes or strong geometric parallax are dynamically emphasized, while degenerate or noise-corrupted views are suppressed before temporal propagation.

With linear projections $\mathbf q_t=\mathbf W_Q\mathbf h_t^I$, $\mathbf k_t^{(i)}=\mathbf W_K\mathbf f_t^{(i)}$, and $\mathbf v_t^{(i)}=\mathbf W_V\mathbf f_t^{(i)}$, cross-attention weights across the six directional tokens are given by:
\begin{equation}
\alpha_t^{(i)}=\frac{\exp(\mathbf q_t^\top\mathbf k_t^{(i)}/\sqrt{d_{\mathrm{head}}})}{\sum_{j=1}^{6}\exp(\mathbf q_t^\top\mathbf k_t^{(j)}/\sqrt{d_{\mathrm{head}}})},
\label{eq:direction_weight}
\end{equation}
evaluated per attention head with head dimension $d_{\mathrm{head}}$. Stacking the directional keys and values into $\mathbf K_t$ and $\mathbf V_t$, multi-head attention is computed with a residual connection and layer normalization:
\begin{equation}
\mathbf z_t=\operatorname{LN}\big(\mathbf q_t+\operatorname{MHA}(\mathbf q_t,\mathbf K_t,\mathbf V_t)\big)\in\mathbb R^{d_{\mathrm{f}}}.
\label{eq:fusion}
\end{equation}
Pixel-level gating suppresses invalid range differences, while cross-attention reweights directional features according to the current platform motion. The residual connection ensures an uninterrupted inertial pathway if exteroceptive range observations become degenerate across all directions.

\subsection{Streaming Causal Transformer with Bounded Memory}
\label{subsec:memory}

Online state estimation on physical robots demands strictly autoregressive inference aligned with chronological sensor telemetry, preventing non-causal future information leakage~\cite{kurt2024causalvio}. While recurrent architectures (e.g., GRUs) naturally enforce causality, their single hidden state creates an information bottleneck that suffers from gradient degradation and memory fading during prolonged maneuvers. Conversely, Transformers capture expressive multi-time-scale kinematic dependencies, but standard self-attention incurs quadratic complexity and acausal dependencies unless strictly constrained. To reconcile causal online execution with long-horizon temporal modeling under severe SWaP-C constraints, we formulate the temporal backbone as a Streaming Causal Transformer governed by a two-tier bounded memory architecture.

In continuous flight, retaining all past key--value (KV) pairs causes cache memory and per-step attention cost to grow linearly with the number of processed frames, increasing pressure on embedded memory and latency budgets. Sliding-window truncation circumvents memory growth but entirely discards distant trajectory history. We resolve this trade-off by decoupling temporal resolution: recent key--value pairs are retained without compression to track agile accelerations, while evicted tokens are progressively compressed into fixed-capacity Chunk-FIFO memory.

Each Transformer layer maintains an uncompressed Local KV cache $\mathcal L_t$ of capacity $C$ for recent key--value pairs. When $\mathcal L_t$ reaches capacity, the oldest $G$ pairs are evicted, pooled, and transformed by a lightweight projection MLP $\phi_m$:
\begin{equation}
[\bar{\mathbf k}_r;\bar{\mathbf v}_r]=\phi_m\left(\frac{1}{G}\sum_{j=1}^{G}[\mathbf k_{r,j};\mathbf v_{r,j}]\right).
\label{eq:chunkkv}
\end{equation}
The resulting summary token is enqueued into Chunk-FIFO memory $\mathcal M_t$ of capacity $S$, discarding the oldest summary upon overflow. The causal attention mechanism then queries the concatenated context:
\begin{equation}
\mathcal C_t=\mathcal L_t\cup\mathcal M_t,\qquad |\mathcal L_t|\le C,\quad|\mathcal M_t|\le S,
\label{eq:memory_bank}
\end{equation}
where lower-triangular causal masking ensures that the query token at step $t$ attends solely to preceding historical states, strictly preventing lookahead bias.

With hyperparameter settings $C=8$, $G=4$, and $S=4$, this memory design spans $24$ frames of temporal history ($\approx1.6$~s at $15$~Hz) while attending to only $12$ retrieval tokens per step. Because the cache capacities and model dimensions are fixed, per-step inference cost and memory footprint remain bounded independently of flight duration. Finally, a lightweight MLP pose head maps the contextualized state to incremental 6-DoF motion estimates $(\Delta\hat{\mathbf p}_t,\Delta\hat{\bm\phi}_t)$.

\subsection{Trajectory Supervision}
\label{subsec:loss}

To mitigate drift accumulated during pose integration, we employ multi-horizon trajectory supervision~\cite{zhao2025tartanimu}, coupling step-wise pose regression with multi-scale manifold constraints, global path-length regularization, and residual smoothness.

Let $\operatorname{SL1}_\beta$ denote the component-averaged Smooth L1 loss with threshold $\beta$. Defining the step errors as $\mathbf e_t^p=\Delta\hat{\mathbf p}_t-\Delta\mathbf p_t$ and $\mathbf e_t^\phi=\Delta\hat{\bm\phi}_t-[\log(\Delta\mathbf R_t)]^\vee$ with rotation vector mapping $[\cdot]^\vee$, the local regression objective is formulated as:
\begin{equation}
\mathcal L_{\mathrm{step}}=\operatorname*{mean}_t\big[\operatorname{SL1}_{0.05}(\mathbf e_t^p)+\operatorname{SL1}_{0.02}(\mathbf e_t^\phi)\big].
\end{equation}

Because small translational biases and orientation drifts compound rapidly, multi-horizon objectives supervise composed motion across horizons $\mathcal H_p=\{4,8,16,32\}$ and $\mathcal H_R=\{4,8,16\}$:
\begin{equation}
\begin{aligned}
\mathcal L_{\mathrm{mh}}^p&=\operatorname*{mean}_{h\in\mathcal H_p,t}\operatorname{SL1}_{0.05}\big((\hat{\mathbf p}_{t+h}-\hat{\mathbf p}_t)-(\mathbf p_{t+h}-\mathbf p_t)\big),\\
\mathcal L_{\mathrm{mh}}^R&=\operatorname*{mean}_{h\in\mathcal H_R,t}\big\|[\log(\Delta\hat{\mathbf R}_{t,h}^{\top}\Delta\mathbf R_{t,h})]^\vee\big\|_2,
\end{aligned}
\end{equation}
where integrated positions $\hat{\mathbf p}_t=\hat{\mathbf p}_0+\sum_{\tau=1}^t\hat{\mathbf R}_{\tau-1}\Delta\hat{\mathbf p}_\tau$ accumulate body-frame translations rotated into the world frame, $\Delta\mathbf R_{t,h}=\Delta\mathbf R_{t+1}\cdots\Delta\mathbf R_{t+h}$ denotes chronological $SO(3)$ composition, and $\mathcal L_{\mathrm{mh}}^R$ penalizes the canonical $SO(3)$ geodesic distance. Errors are averaged uniformly across valid horizons.

To constrain global scale fidelity and suppress high-frequency prediction jitter, we incorporate trajectory scale and residual smoothness objectives:
\begin{equation}
\begin{aligned}
\mathcal L_{\mathrm{scale}}&=\operatorname{SL1}_{0.10}\left(\frac{\sum_t\|\Delta\hat{\mathbf p}_t\|_2}{\max(\sum_t\|\Delta\mathbf p_t\|_2,10^{-6})}-1\right),\\
\mathcal L_{\mathrm{res}}&=\operatorname*{mean}_{t\ge2}\operatorname{SL1}_{0.02}(\mathbf e_t^p-\mathbf e_{t-1}^p).
\end{aligned}
\end{equation}
$\mathcal L_{\mathrm{scale}}$ penalizes path-length bias, while $\mathcal L_{\mathrm{res}}$ penalizes step-to-step changes in translation error, reducing high-frequency prediction jitter without directly penalizing rapid vehicle motion. The joint training objective is:
\begin{equation}
\mathcal L=\mathcal L_{\mathrm{step}}+0.02\mathcal L_{\mathrm{mh}}^p+0.005(\mathcal L_{\mathrm{mh}}^R+\mathcal L_{\mathrm{scale}}+\mathcal L_{\mathrm{res}}).
\end{equation}

%% file: sections_en/experiments.tex
\section{Experiments and Results}
\label{sec:experiments}

We evaluate tracking accuracy, multi-environment performance, robustness under severe sensing degradation, individual architectural components, and onboard real-time execution.

\subsection{Experimental Setup and Evaluation Protocol}
\label{subsec:setup}

\begin{figure}[!ht]
\centering
\includegraphics[width=0.96\columnwidth,trim=0 120bp 0 96bp,clip]{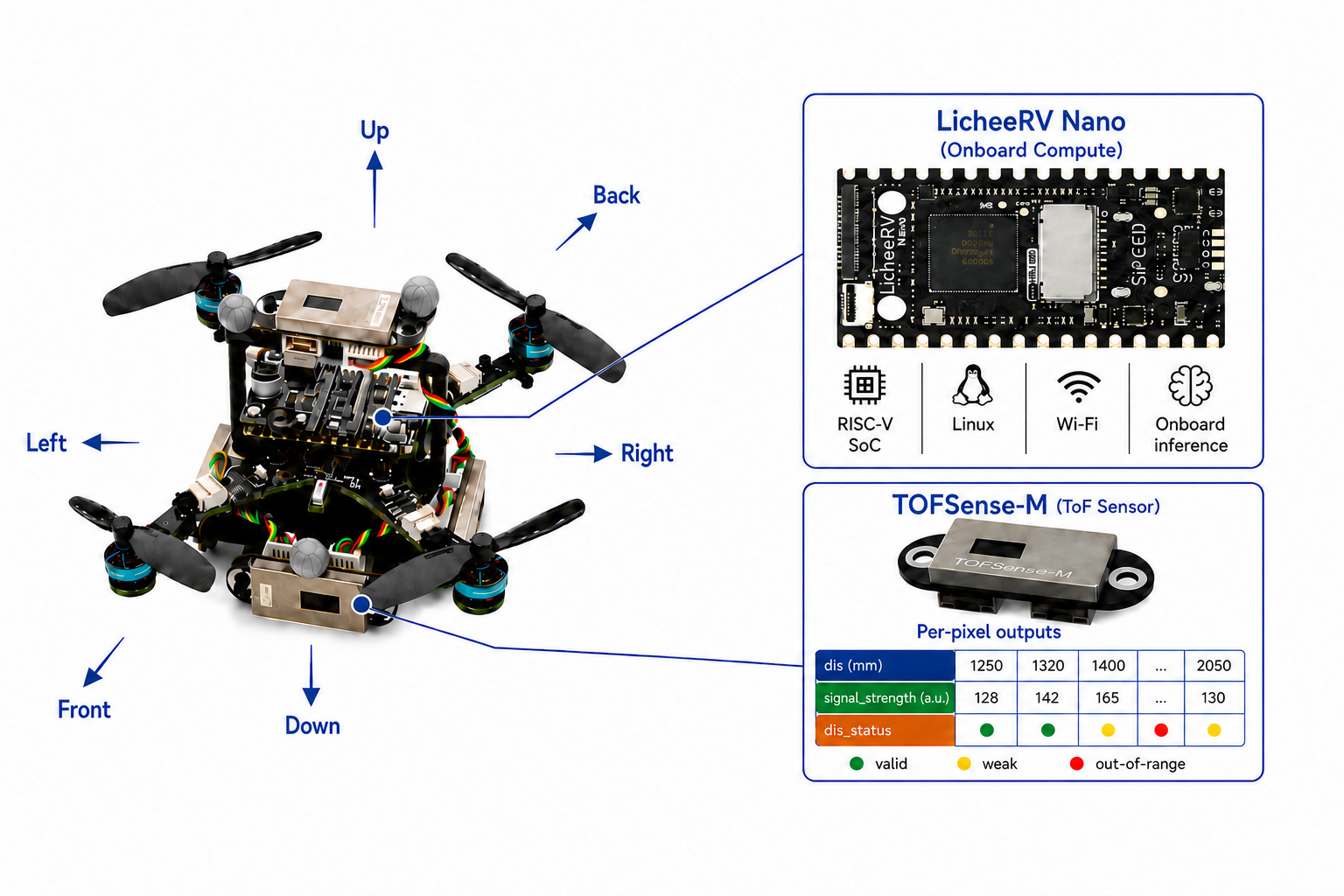}
\caption{Crazyflie nano-UAV platform equipped with six orthogonal TOFSense-M arrays, an onboard 200~Hz IMU, and a LicheeRV Nano companion computer for time-synchronized streaming inference.}
\label{fig:hardware_platform}
\end{figure}

\textbf{1) Flight Benchmark (Crazyflie Platform):}
Flight experiments were conducted on a Crazyflie nano-UAV equipped with a custom sensor expansion deck (Fig.~\ref{fig:hardware_platform}). The onboard companion computer (LicheeRV Nano with an SG2002 SoC) samples the six-axis IMU via CPX at $200$~Hz and queries six TOFSense-M sensors over UART to acquire orthogonal $8\times8$ range grids at $15$~Hz. Each sensor module has a mass of $2.5$~g, contributing an aggregate exteroceptive payload of only $15$~g. Ground-truth poses were recorded via an external motion capture system across both canonical geometric trajectories (circles, figure-eights, squares, straight lines) and random 3D flights. The benchmark comprises $9{,}492.01$~s of flight across $2{,}195.22$~m of total trajectory length, with canonical paths accounting for $20\%$ of the dataset and random flights constituting the remaining $80\%$. Random flights are partitioned into training, validation, and test sets ($70:15:15$); all canonical paths and held-out test sequences (standardized to $4$--$7$~m trajectory lengths) are strictly reserved for testing.

To ensure multi-sensor temporal synchronization, IMU timestamps are mapped to the SG2002 monotonic clock using four-timestamp bidirectional clock synchronization between the flight controller and companion computer, aligning inertial telemetry with the multi-view ToF acquisitions. Network inputs adhere to a fixed spatial order: front, right, back, down, left, and up.

\textbf{2) Multi-Environment Dataset:}
To assess performance across diverse spatial geometries, the identical sensing payload was deployed on an extended-endurance aerial platform across four distinct real-world scenes spanning $2{,}362$~m of active flight: multiple office spaces ($1{,}368$~m, dense non-convex clutter), a pantry ($524$~m, reflective tiles and specular surfaces), a long corridor ($194$~m, longitudinal geometric degeneracy), and a meeting room ($276$~m). Each environment is partitioned under the identical $70:15:15$ protocol with test trajectories ranging from $4$ to $7$~m, providing a standardized benchmark for multi-environment evaluation.

\textbf{3) Baselines and Evaluation Metrics:}
We benchmark \ours{} against three representative paradigms: 1) \textbf{Crazyflie Flow}, the commercial baseline combining downward optical flow, single-point ToF, and complementary filtering; 2) learning-based inertial odometry baselines \textbf{TLIO~\cite{liu2020tlio}} and \textbf{AirIO~\cite{qiu2025airio}}; and 3) \textbf{Naive ToF+IMU}, an ablative baseline processing identical multi-view ToF and IMU streams via static channel concatenation without cross-attention or temporal memory.

\begin{figure}[!ht]
\centering
{\scriptsize
\textcolor[HTML]{111111}{\rule{8pt}{1.2pt}}~GT\quad
\textcolor[HTML]{D62728}{\rule{8pt}{1.2pt}}~Ours\quad
\textcolor[HTML]{9467BD}{\rule{8pt}{1.2pt}}~Flow\quad
\textcolor[HTML]{1F77B4}{\rule{8pt}{1.2pt}}~AirIO\quad
\textcolor[HTML]{FF7F0E}{\rule{8pt}{1.2pt}}~TLIO}
\par\vspace{0.4mm}
\includegraphics[width=0.325\columnwidth]{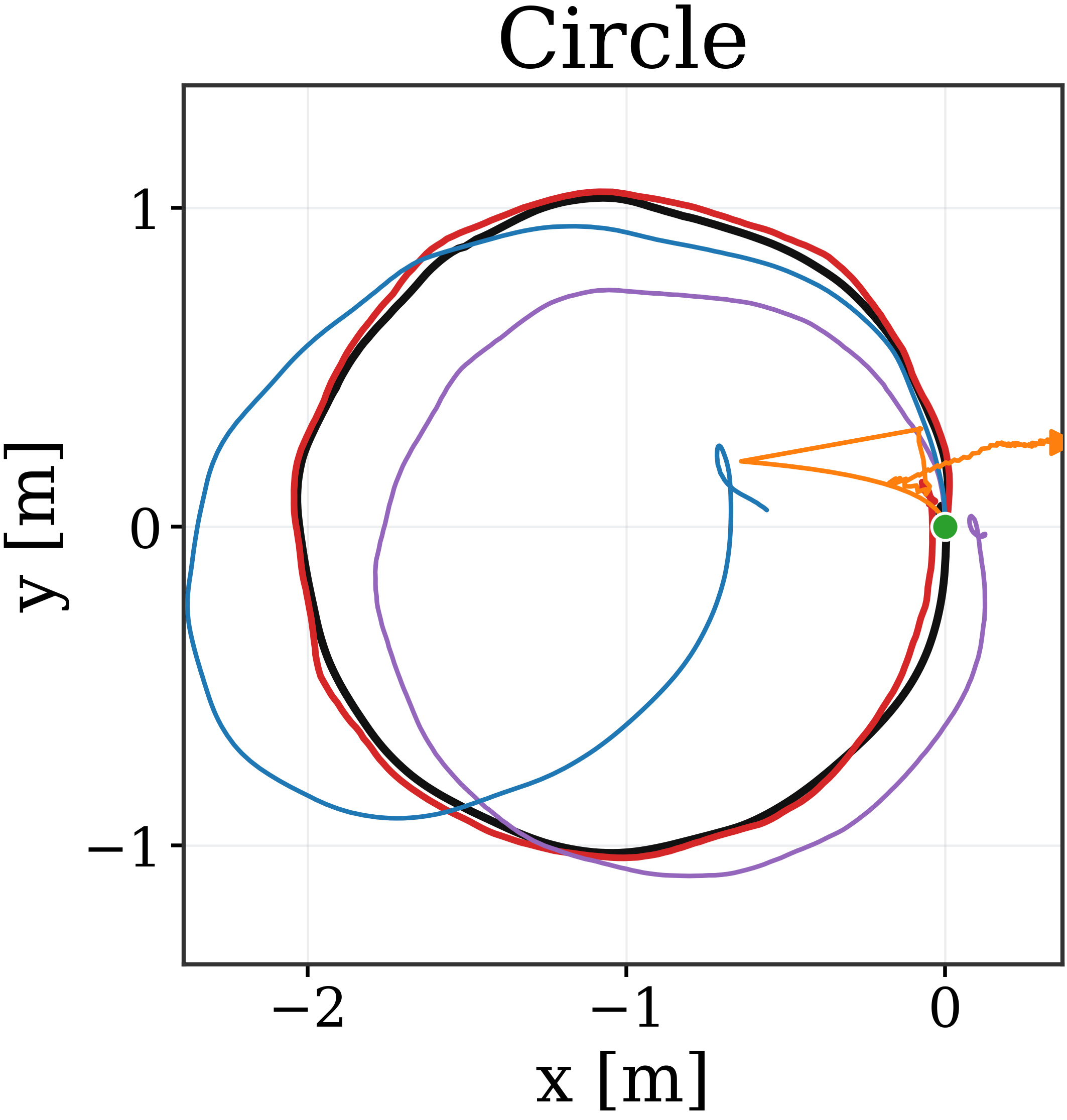}\hfill
\includegraphics[width=0.325\columnwidth]{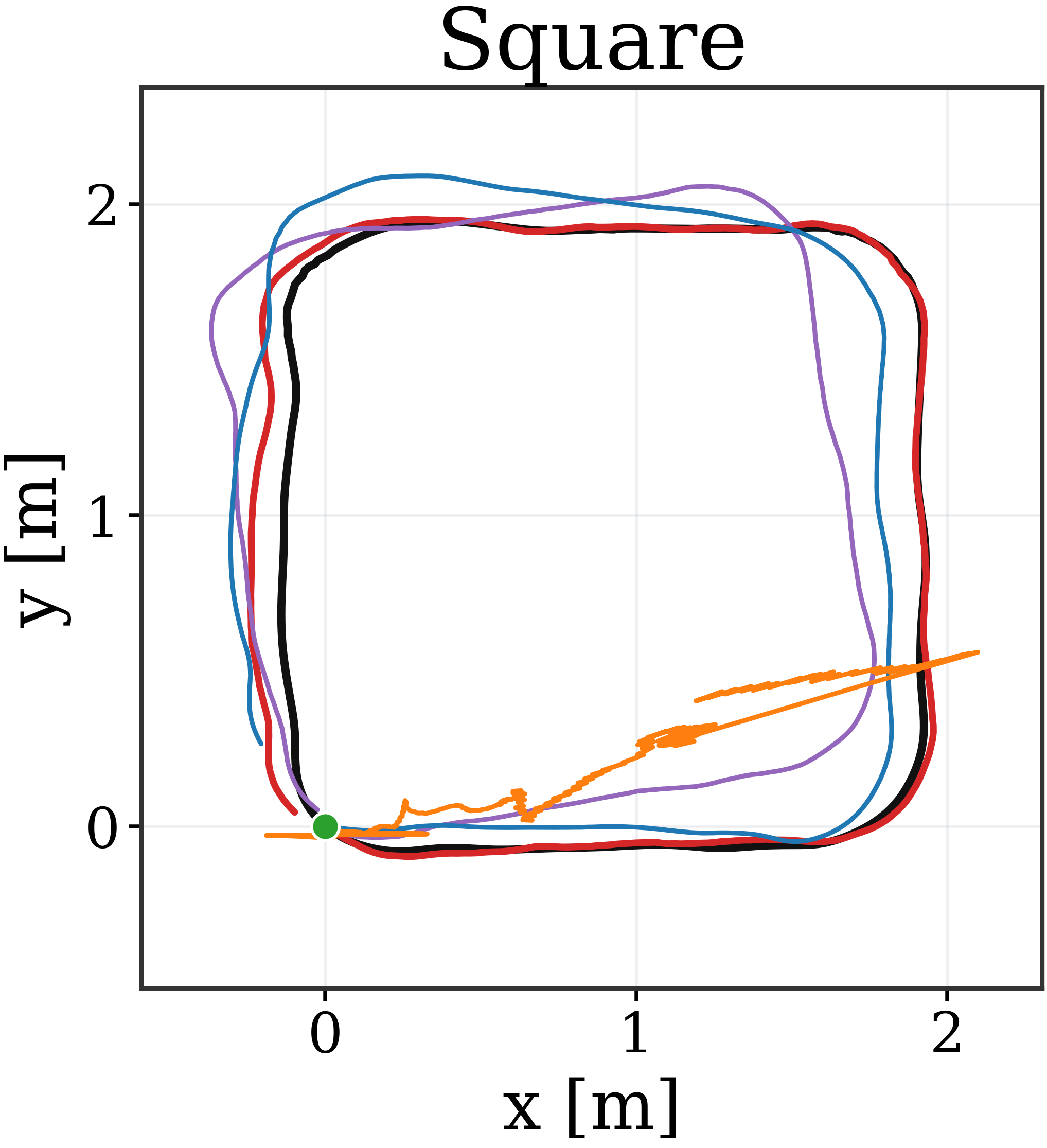}\hfill
\includegraphics[width=0.325\columnwidth]{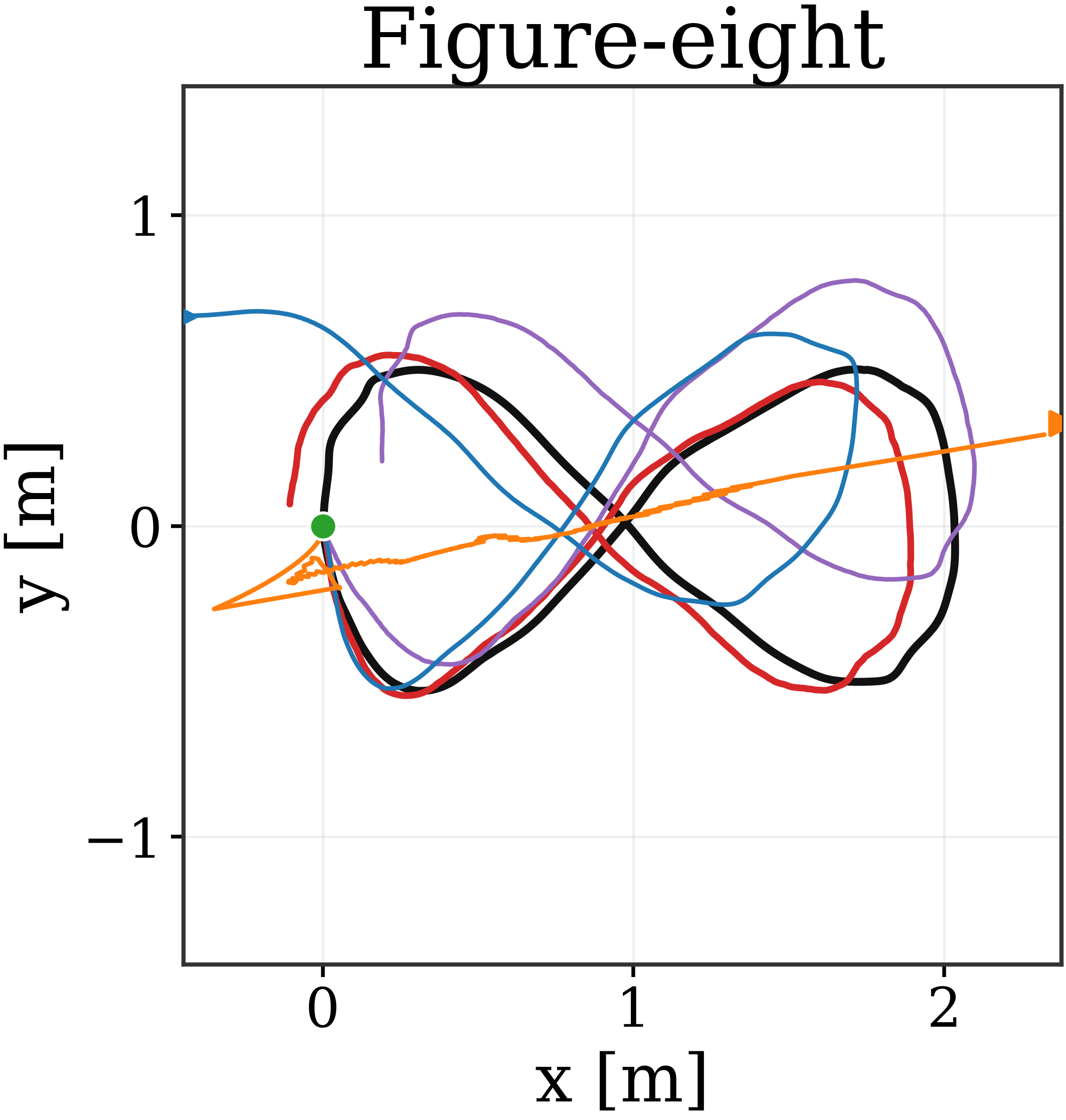}
\par\vspace{0.5mm}
\includegraphics[width=0.325\columnwidth]{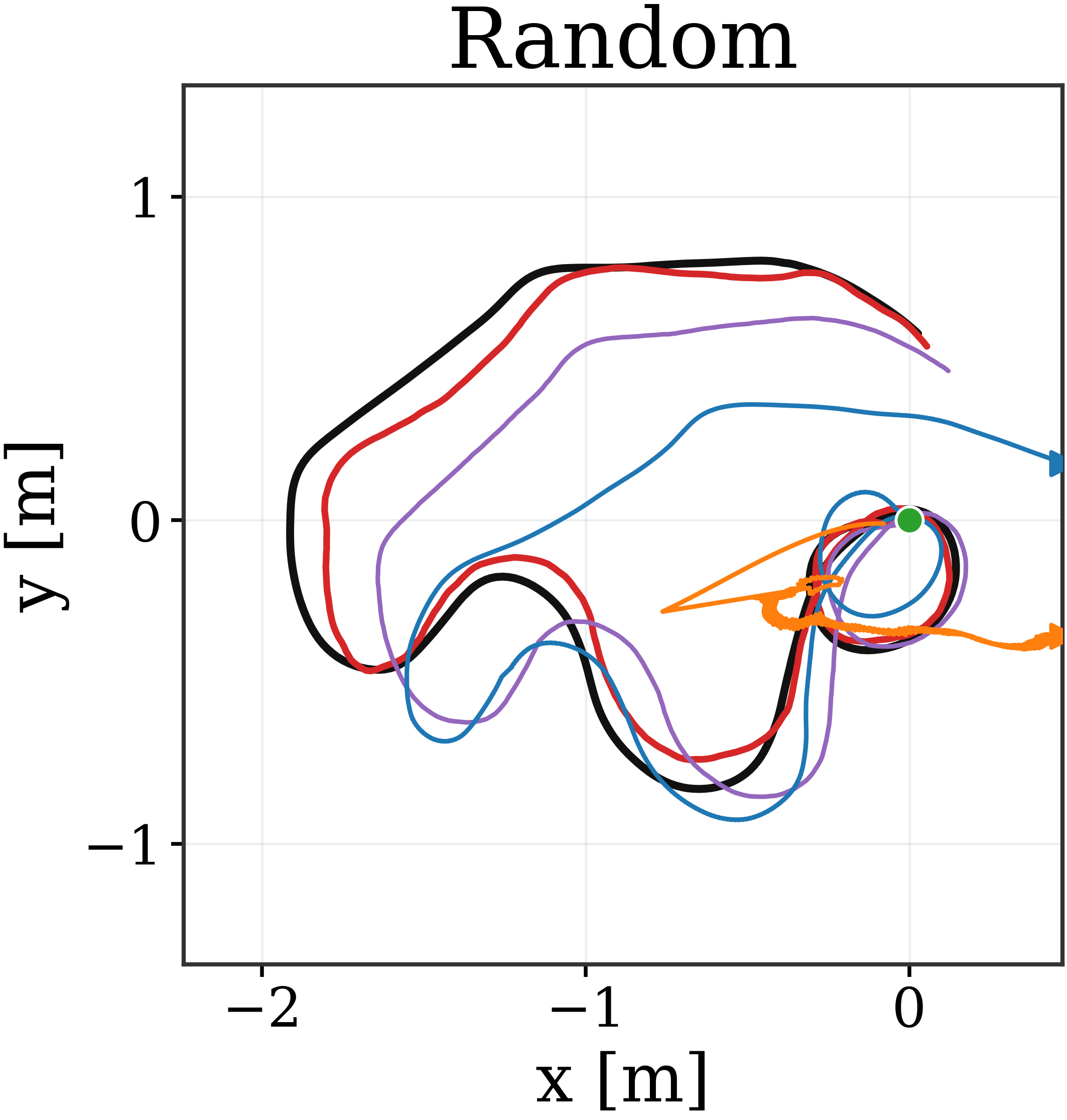}\hfill
\includegraphics[width=0.325\columnwidth]{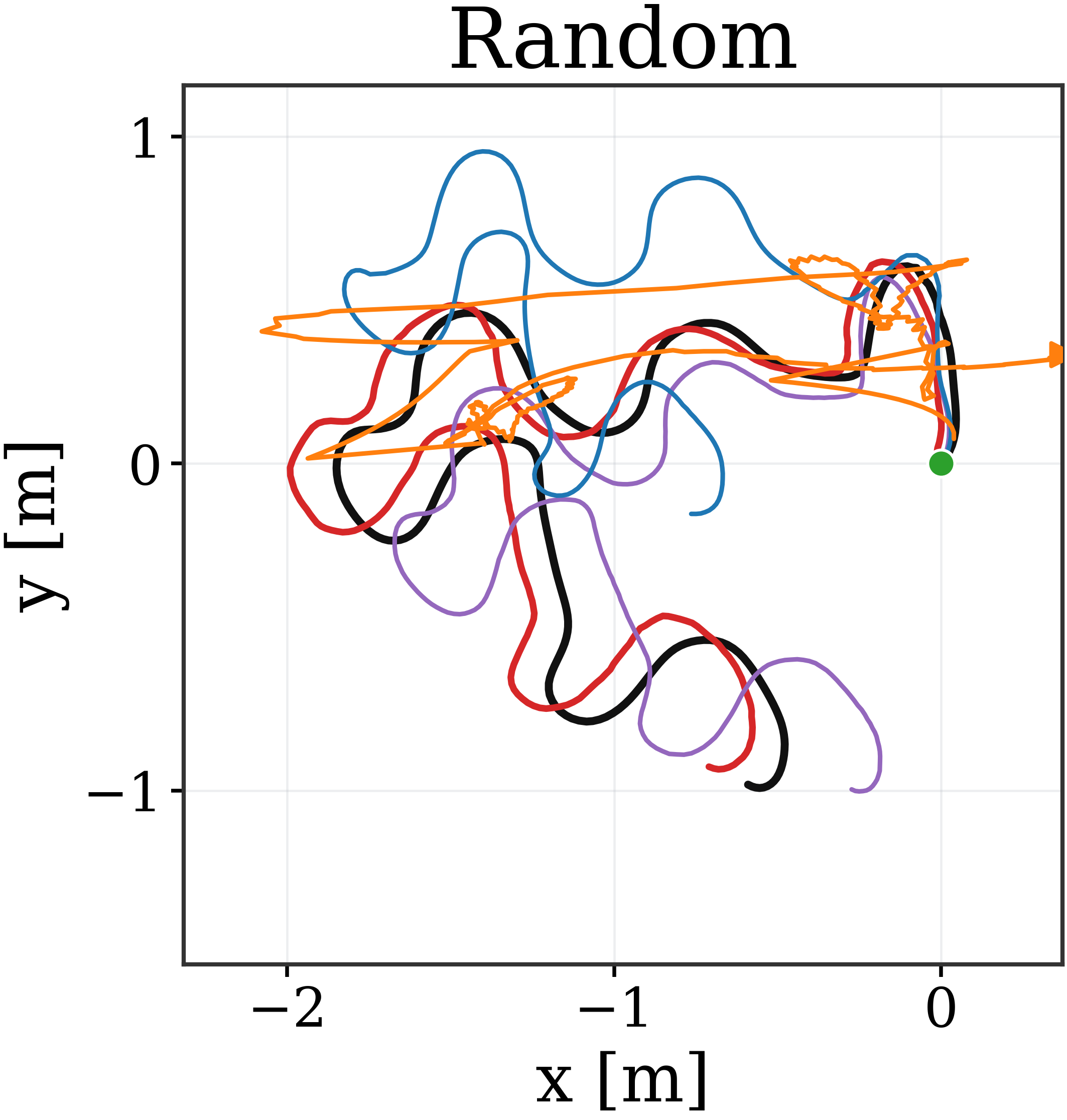}\hfill
\includegraphics[width=0.325\columnwidth]{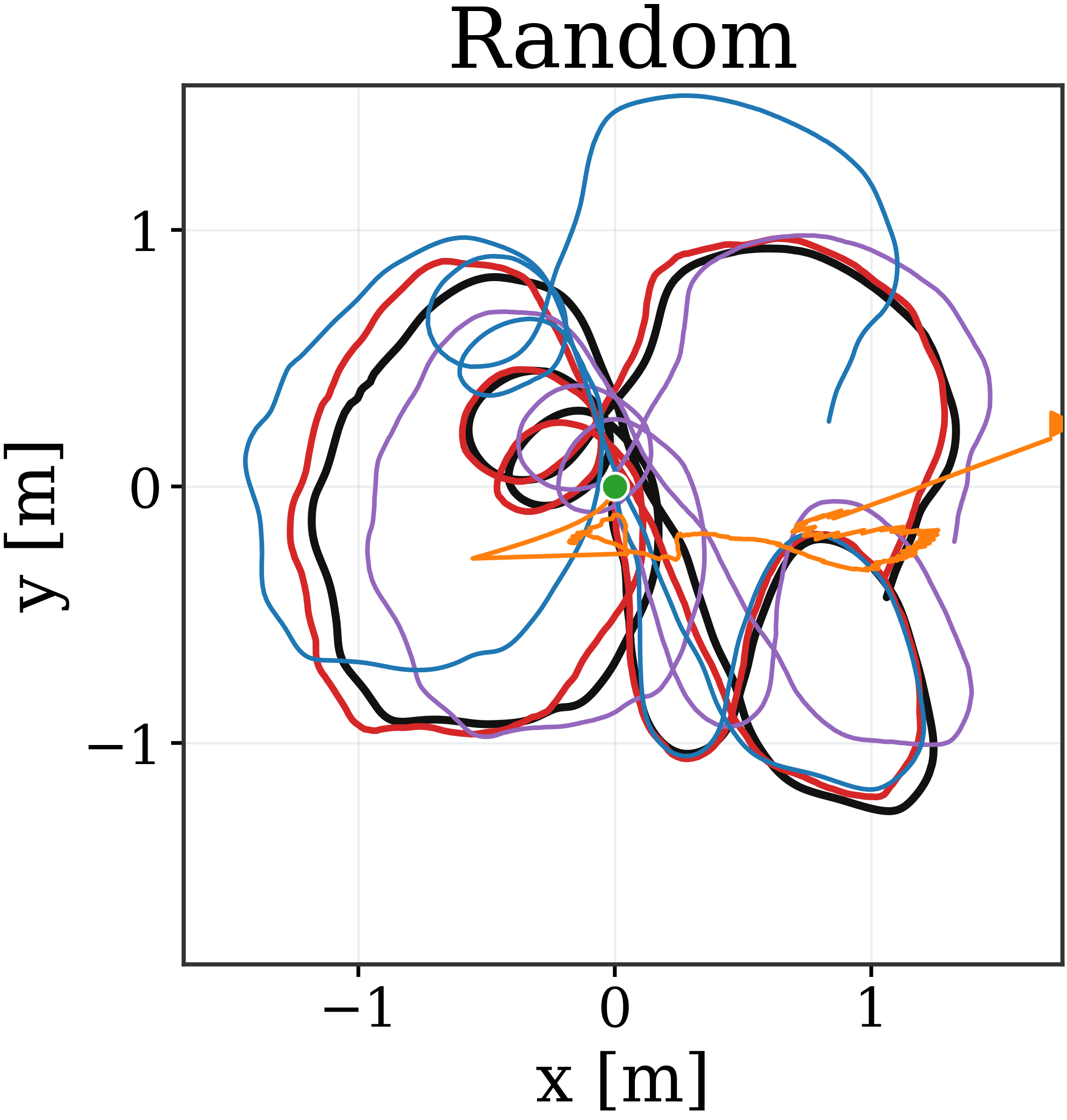}
\caption{Qualitative dead-reckoning trajectory comparisons across canonical paths (circle, square, figure-eight) and random flights. All trajectories represent unaligned open-loop integrations from the initial pose.}
\label{fig:trajectory_comparison}
\end{figure}

Tracking accuracy is quantified using position Absolute Trajectory Error ($\mathrm{ATE}_p$), frame-to-frame relative translation and rotation errors ($\mathrm{RPE}_p$, $\mathrm{RPE}_R$ at $66.7$~ms), and Endpoint Drift Rate $\mathrm{EDR}=\|\hat{\mathbf p}_T-\mathbf p_T\|_2/L_{\mathrm{gt}}\times100\%$. ATE and RPE are computed as per-sequence RMSE and averaged over the test set. For learning-based models, tables report the mean and standard deviation over three random seeds.

\subsection{System-Level Tracking and Multi-Environment Evaluation}
\label{subsec:system_tracking}

\begin{table}[!t]
\centering
\caption{Open-loop tracking accuracy on the test set.}
\label{tab:main}
\scriptsize
\setlength{\tabcolsep}{1.0pt}
\resizebox{0.98\columnwidth}{!}{%
\begin{tabular}{@{}lcccc@{}}
\toprule
Method / Input & \shortstack{ATE$_p$\\(m)$\downarrow$} & \shortstack{RPE$_p$\\(m)$\downarrow$} & \shortstack{RPE$_R$\\(deg)$\downarrow$} & \shortstack{EDR\\(\%)$\downarrow$}\\
\midrule
\shortstack[l]{Crazyflie Flow} & $0.259$ & $0.0079$ & $0.576$ & $11.65$\\
TLIO~\cite{liu2020tlio} (IMU) & $1.084{\pm}0.011$ & $0.0519{\pm}0.0002$ & $0.603{\pm}0.001$ & $57.46{\pm}0.32$\\
AirIO~\cite{qiu2025airio} (IMU) & $0.351{\pm}0.050$ & $0.0067{\pm}0.0007$ & $0.571{\pm}0.002$ & $24.49{\pm}3.49$\\
\midrule
Naive ToF+IMU & $0.462{\pm}0.075$ & $0.0069{\pm}0.0004$ & $0.532{\pm}0.009$ & $28.46{\pm}3.47$\\
\rowcolor{black!6}
\textbf{\ours{} (Ours)} & $\mathbf{0.118{\pm}0.007}$ & $\mathbf{0.0039{\pm}0.0001}$ & $\mathbf{0.504{\pm}0.001}$ & $\mathbf{7.18{\pm}0.37}$\\
\bottomrule
\end{tabular}
}
\end{table}

Table~\ref{tab:main} shows that \ours{} achieves $0.118$~m position ATE, reducing error by $54.4\%$ relative to Crazyflie Flow and by $66.4\%$--$89.1\%$ relative to the learned inertial baselines. Lower endpoint drift supports the role of sparse range measurements as external geometric constraints on inertial integration.

The higher error of Naive ToF+IMU suggests that directional fusion and temporal context help exploit range geometry beyond simply adding sensor inputs. Fig.~\ref{fig:trajectory_comparison} shows better preservation of loops and sharp corners with \ours{}.

\textbf{Multi-Environment Performance:}
Table~\ref{tab:generalization} reports the lowest ATE for \ours{} in all four scenes ($0.091$--$0.158$~m). These results establish performance on held-out sequences within each environment. The lowest error in the pantry may reflect stronger geometric constraints from nearby surfaces. Crazyflie Flow is omitted because of frequent tracking divergence on low-texture surfaces.

\begin{table}[!t]
\centering
\caption{Multi-environment accuracy ($\mathrm{ATE}_p$, m; mean$\pm$std over three seeds).}
\label{tab:generalization}
\setlength{\tabcolsep}{2pt}
\scriptsize
\begin{tabular*}{\columnwidth}{@{\extracolsep{\fill}}lcccc@{}}
\toprule
Method & Corridor & Office & Pantry & Meeting room\\
\midrule
TLIO~\cite{liu2020tlio} (IMU) & $1.241{\pm}0.038$ & $0.985{\pm}0.021$ & $1.152{\pm}0.045$ & $1.026{\pm}0.019$ \\
AirIO~\cite{qiu2025airio} (IMU) & $0.412{\pm}0.035$ & $0.345{\pm}0.028$ & $0.389{\pm}0.032$ & $0.356{\pm}0.024$ \\
\midrule
Naive ToF+IMU & $0.531{\pm}0.082$ & $0.418{\pm}0.063$ & $0.412{\pm}0.091$ & $0.443{\pm}0.055$ \\
\rowcolor{black!6}
\textbf{Ours} & $\mathbf{0.142{\pm}0.009}$ & $\mathbf{0.158{\pm}0.012}$ & $\mathbf{0.091{\pm}0.006}$ & $\mathbf{0.114{\pm}0.007}$ \\
\bottomrule
\end{tabular*}
\end{table}

\subsection{Robustness Under Directional Blind Spots and Sensing Degeneracy}
\label{subsec:robustness}

Physical deployments frequently encounter sensor dropouts, structural openings, and non-reflective surfaces. We evaluate robustness under these sensing conditions.

\begin{figure}[!ht]
\centering
\includegraphics[width=0.96\columnwidth]{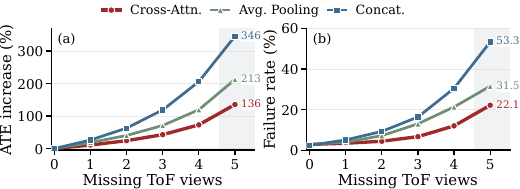}
\caption{Fusion robustness under directional observation degradation. (a) Relative $\mathrm{ATE}_p$ degradation; (b) failure rate ($\mathrm{EDR}>20\%$). Curves aggregate persistent, random, and burst failures; shading marks the extreme five-view-loss setting.}
\label{fig:directional_degradation}
\end{figure}

\textbf{1) Dynamic Routing Robustness under Directional Loss:}
Under nominal sensing, cross-attention reduces position ATE by $7.8\%$ relative to concatenation ($0.118$~m vs. $0.128$~m). Its advantage broadens substantially under directional view dropouts (Fig.~\ref{fig:directional_degradation}): across one to five missing views, the average relative ATE increase is restricted to $57.19\%$ for cross-attention, compared to $151.84\%$ for concatenation. When five of the six directional sensors are masked out, cross-attention retains an ATE of $0.307$~m and a failure rate of $22.12\%$, whereas concatenation yields an ATE of $0.569$~m and a failure rate of $53.31\%$. These results demonstrate that motion-conditioned cross-attention maintains lower trajectory error and failure rates than concatenation under directional observation loss.

\begin{figure}[!t]
\centering
\includegraphics[width=0.96\columnwidth]{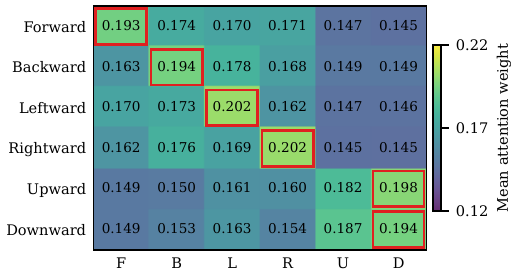}
\caption{Directional cross-attention weights conditioned on dominant translational kinematics, evaluated on the test set (mean over three seeds). Red boxes highlight row-wise maxima. F, B, L, R, U, and D denote front, back, left, right, up, and down directions, respectively.}
\label{fig:attention_motion}
\end{figure}

\textbf{Motion-Conditioned Attention Visualization:}
To interpret the learned behavior of the IMU-guided cross-attention module, we analyze the directional attention distributions $\alpha_t^{(i)}$ conditioned on dominant body-frame translational kinematics across the test set (Fig.~\ref{fig:attention_motion}). Flight intervals are grouped by six translation directions (forward, backward, leftward, rightward, upward, and downward) using ground-truth velocity, with attention weights averaged over all heads, temporal intervals, and three random seeds.

As shown in Fig.~\ref{fig:attention_motion}, the routing module exhibits pronounced motion-correlated directional selectivity. During horizontal motion, IMU-guided attention assigns the highest average weights to views aligned with the motion direction, producing the diagonal pattern across F, B, L, and R. In contrast, during vertical maneuvers (both upward and downward), the network consistently concentrates the highest attention on the downward sensor. This preference may be explained by more consistent range returns from nearby floors, whereas upward measurements can be affected by ceiling distance and fixtures.

\begin{figure}[!ht]
\centering
\includegraphics[width=0.96\columnwidth]{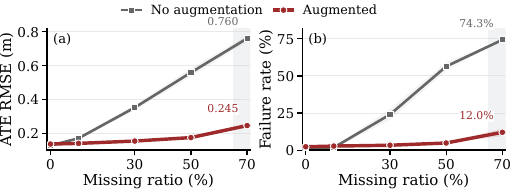}
\caption{Ablation of controlled degradation augmentation under varying missing rates. (a) $\mathrm{ATE}_p$ RMSE; (b) failure rate ($\mathrm{EDR}>20\%$). Curves and uncertainty bands denote mean and standard deviation across three degradation modes; background shading marks the extreme $\rho=70\%$ out-of-distribution (OOD) condition withheld during training.}
\label{fig:controlled_missing}
\end{figure}

\textbf{2) Resilience to Measurement Corruption via Stochastic Augmentation:}
Fig.~\ref{fig:controlled_missing} compares training with and without stochastic degradation augmentation. Under missing pixel ratios of $30\%$ and $50\%$, the augmented model maintains low ATEs of $0.154$~m and $0.174$~m, respectively. Even under extreme out-of-distribution corruption with $70\%$ missing measurements, the augmented model maintains an ATE of $0.245$~m with a $12.0\%$ failure rate, whereas the unaugmented baseline diverges rapidly ($0.760$~m ATE, $74.3\%$ failure rate).

\subsection{Ablation Study on Core Architectural Mechanisms}
\label{subsec:ablation}

Ablations on the benchmark test set isolate the contributions of individual subsystems (Tables~\ref{tab:comprehensive_ablation} and~\ref{tab:temporal_ablation}).

\begin{table}[!ht]
\centering
\caption{Component ablations on the test set.}
\label{tab:comprehensive_ablation}
\scriptsize
\setlength{\tabcolsep}{2.5pt}
\begin{tabular}{@{}l>{\raggedright\arraybackslash}p{0.37\columnwidth}cc@{}}
\toprule
Component & Variant & \shortstack{ATE$_p$\\(m)$\downarrow$} & \shortstack{EDR\\(\%)$\downarrow$}\\
\midrule
\rowcolor{black!6}
Reference & \textbf{TIO-Former (Ours)} & $\mathbf{0.118{\pm}0.007}$ & $\mathbf{7.18{\pm}0.37}$\\
\midrule
\multirow{2}{*}{\shortstack[l]{Sensor\\Modality}}
& IMU-only baseline & $0.816{\pm}0.023$ & $39.48{\pm}1.53$\\
& ToF-only baseline & $0.156{\pm}0.009$ & $8.76{\pm}0.13$\\
\midrule
\multirow{2}{*}{\shortstack[l]{Spatial\\Coverage}}
& Down-view only & $0.937{\pm}0.003$ & $47.52{\pm}1.92$\\
& Four lateral views only & $0.129{\pm}0.002$ & $7.29{\pm}0.28$\\
\midrule
\multirow{2}{*}{\shortstack[l]{Range\\Rep.}}
& Single frame $\range_t$ & $0.306{\pm}0.004$ & $14.29{\pm}0.58$\\
& Metric pair $[\range_{t-1},\range_t]$ & $0.158{\pm}0.004$ & $8.78{\pm}0.73$\\
\midrule
\multirow{2}{*}{\shortstack[l]{Cross\\Fusion}}
& Channel Concatenation & $0.128{\pm}0.002$ & $7.82{\pm}0.25$\\
& Masked Average Pooling & $0.126{\pm}0.007$ & $7.46{\pm}0.31$\\
\midrule
\multirow{1}{*}{View Enc.}
& W/o View Transformer & $0.129{\pm}0.003$ & $7.25{\pm}0.71$\\
\midrule
\multirow{2}{*}{\shortstack[l]{Loss \&\\Training}}
& Single-step $\mathcal L_{\mathrm{step}}$ only & $0.159{\pm}0.005$ & $9.39{\pm}0.72$\\
& W/o $\mathcal L_{\mathrm{scale}}$ \& $\mathcal L_{\mathrm{res}}$ terms & $0.132{\pm}0.004$ & $8.12{\pm}0.41$\\
\bottomrule
\end{tabular}
\end{table}

\textbf{1) Sensor Modality and Spatial View Coverage:}
Table~\ref{tab:comprehensive_ablation} shows that both modalities contribute, with a larger penalty when ToF is removed. Four lateral views approach six-view accuracy, whereas downward-only sensing produces severe drift. Over planar ground, downward ranges mainly constrain altitude and tilt; lateral views provide additional constraints on horizontal motion.

\textbf{2) Causal Imputation and Bilateral Gated Difference:}
Building on the metric pair, causal imputation and the bilateral gated difference reduce ATE by $25.3\%$. Imputation fills missing entries while preserving observed metric ranges, and the gated difference provides explicit range-change cues while suppressing artifacts from validity transitions.

\textbf{3) Multi-Horizon Trajectory Supervision:}
Multi-horizon supervision improves ATE by $25.8\%$ over single-step training. Scale and residual regularization further improve accuracy: the former penalizes path-length bias, while the latter discourages abrupt changes in prediction error. Together with multi-horizon supervision, they target accumulated drift and local jitter.

\begin{table}[t]
\centering
\caption{Temporal memory: accuracy and onboard execution.}
\label{tab:temporal_ablation}
\scriptsize
\setlength{\tabcolsep}{2pt}
\textit{(a) Trajectory accuracy}\par\vspace{1pt}
\begin{tabular*}{\columnwidth}{@{\extracolsep{\fill}}lccc@{}}
\toprule
Backbone & Tok./Hist. & \shortstack{ATE$_p$\\(m)$\downarrow$} & \shortstack{EDR\\(\%)$\downarrow$}\\
\midrule
Memoryless & 0 / 1 & $0.525{\pm}0.125$ & $32.04{\pm}6.32$\\
GRU2 & --- & $0.206{\pm}0.083$ & $12.28{\pm}5.41$\\
KV8 & 8 / 8 & $0.134{\pm}0.008$ & $7.39{\pm}0.12$\\
KV12 & 12 / 12 & $0.134{\pm}0.002$ & $7.58{\pm}0.36$\\
KV24 & 24 / 24 & $0.133{\pm}0.006$ & $7.59{\pm}0.23$\\
KV32 & 32 / 32 & $0.130{\pm}0.007$ & $7.36{\pm}0.25$\\
KV8+FIFO4 (Avg.) & 12 / 24 & $0.129{\pm}0.006$ & $7.35{\pm}0.38$\\
\rowcolor{black!6}
\textbf{KV8+FIFO4 (Ours)} & \textbf{12 / 24} & $\mathbf{0.118{\pm}0.007}$ & $\mathbf{7.18{\pm}0.37}$\\
\bottomrule
\end{tabular*}
\par\vspace{2pt}
\textit{(b) Reference accuracy and deployment}\par\vspace{1pt}
\begin{tabular*}{\columnwidth}{@{\extracolsep{\fill}}lcccc@{}}
\toprule
Backbone & \shortstack{ATE$_{\rm ref}$\\(m)$\downarrow$} & \shortstack{P95\\(ms)$\downarrow$} & \shortstack{RSS\\(MiB)$\downarrow$} & \shortstack{$\delta_{\max}$\\(mm)$\downarrow$}\\
\midrule
Memoryless & $2.908{\pm}0.709$ & 6.413 & 6.281 & 22.80\\
GRU2 & $1.675{\pm}0.664$ & 7.732 & 4.26 & 19.35\\
KV8 & $0.526{\pm}0.086$ & 8.767 & 6.301 & 25.86\\
KV12 & $0.486{\pm}0.058$ & 8.853 & 6.316 & 10.09\\
KV24 & $0.475{\pm}0.082$ & 9.059 & 6.363 & 10.36\\
KV32 & $0.518{\pm}0.033$ & 9.216 & 6.395 & 10.87\\
KV8+FIFO4 (Avg.) & $0.456{\pm}0.108$ & 9.661 & 5.69 & 17.41\\
\rowcolor{black!6}
\textbf{KV8+FIFO4 (Ours)} & $\mathbf{0.389{\pm}0.013}$ & 10.466 & 6.324 & $\mathbf{5.41}$\\
\bottomrule
\end{tabular*}
\end{table}

\textbf{4) Streaming Temporal Memory Trade-Offs:}
In Table~\ref{tab:temporal_ablation}(a), Tok./Hist. denotes retrieved tokens/history frames, and Avg. denotes unprojected mean pooling. Memoryless estimation accumulates severe drift ($0.525$~m ATE), while GRU2 exhibits greater variability across random seeds. In contrast, linearly expanding the uncompressed KV window from 8 to 32 tokens yields diminishing accuracy gains ($0.134$~m to $0.130$~m) while scaling memory consumption.

Under a fixed 12-token retrieval budget, KV8+FIFO4 extends context to 24 frames and outperforms KV12 and unprojected mean pooling. It also improves accuracy by $11.3\%$ over uncompressed KV24 while halving active token retrieval, showing that compressed history is more useful than a larger uncompressed cache under the real-time deadline.

\textbf{Edge Deployment Consistency:}
Table~\ref{tab:temporal_ablation}(b) evaluates five $20$~m random flights. The maximum cumulative position discrepancy between onboard mixed-precision and Python FP32 outputs is $\delta_{\max}=\max_t\|\hat{\mathbf p}^{\mathrm{board}}_t-\hat{\mathbf p}^{\mathrm{Python}}_t\|_2$. \ours{} achieves the lowest $\delta_{\max}$ ($5.41$~mm) and reference ATE ($0.389{\pm}0.013$~m), confirming accurate long-horizon streaming under mixed-precision deployment.

\subsection{Onboard Real-Time Streaming Deployment on Edge Hardware}
\label{subsec:deployment}

\begin{table}[!ht]
\centering
\caption{Onboard resource use and timing of \ours{}.}
\label{tab:onboard_resources}
\scriptsize
\setlength{\tabcolsep}{4pt}
\begin{tabularx}{\columnwidth}{@{}Xr@{}}
\toprule
Metric & Value\\
\midrule
Embedded Hardware SoC & SG2002 (C906@850 MHz + CV181x TPU)\\
Heterogeneous Pipeline & TPU (BF16) + CPU RVV (FP32)\\
Model Parameters / Storage Footprint & \textbf{1.117 M} / 3.958 MiB\\
Peak Resident Set Size (RSS) & \textbf{6.324 MiB} (4.94\% of 128 MiB RAM)\\
\midrule
Per-Step Latency (Mean / P95) & 10.038 ms / \textbf{10.466 ms}\\
Maximum Observed Latency & 20.609 ms (30.90\% of budget)\\
Single-Core CPU Utilization & 14.89\%\\
Inverse Mean Compute Time & 99.62 Hz\\
Deadline Overruns ($>66.7$ ms) & \textbf{0 / 5,000 (0.000\%)}\\
\bottomrule
\end{tabularx}
\end{table}

On the SG2002, the ToF CNN runs on the TPU in BF16, while IMU encoding, directional fusion, and streaming memory run on the 850~MHz RISC-V CPU using FP32 RVV. Table~\ref{tab:onboard_resources} reports a P95 latency of $10.466$~ms, a peak RSS of $6.324$~MiB, and zero deadline overruns across $5{,}000$ cycles, confirming real-time streaming under the tested conditions.

%% file: sections_en/conclusion.tex
\section{Conclusion}
\label{sec:conclusion}

We presented \ours{}, a camera-free, optical-flow-free, and mapless 6-DoF range-inertial odometry framework for nano-UAVs. The bilateral gated difference, IMU-guided directional fusion, and two-tier temporal memory jointly improve tracking over optical-flow and learned inertial baselines in the evaluated flights. Fixed cache capacities bound per-step inference cost and memory independently of flight duration; onboard measurements demonstrate real-time processing under the tested conditions. Limited sensing range and geometric degeneracy remain challenges for open-loop estimation. Future work will explore lightweight loop closure and self-supervised pretraining to improve long-term robustness.